\documentclass[fleqn,10pt]{wlscirep}
\usepackage[utf8]{inputenc}
\usepackage[T1]{fontenc}
\usepackage[ruled, linesnumbered ]{algorithm2e}
\usepackage{cleveref}
\usepackage{adjustbox}
\usepackage{multirow}
\usepackage{placeins}
\usepackage{xurl}

\title{Feature-Weighted Maximum Representative Subsampling}

\author[1,*]{Tony Hauptmann}
\author[1]{Stefan Kramer}
\affil[1]{Institute of Computer Science, Johannes Gutenberg University Mainz, Mainz, Germany}

\affil[*]{Corresponding author: thauptmann@uni-mainz.de}

\keywords{debiasing, feature weights}

\begin{abstract}
In the social sciences, it is often necessary to debias studies and surveys before valid conclusions can be drawn. Debiasing algorithms enable the computational removal of bias using sample weights. However, an issue arises when only a subset of features is highly biased, while the rest is already representative. Algorithms need to strongly alter the sample distribution to manage a few highly biased features, which can in turn introduce bias into already representative variables. To address this issue, we developed a method that uses feature weights to minimize the impact of highly biased features on the computation of sample weights. Our algorithm is based on Maximum Representative Subsampling (MRS), which debiases datasets by aligning a non-representative sample with a representative one through iterative removal of elements to create a representative subsample. The new algorithm, named feature-weighted MRS (FW-MRS), decreases the emphasis on highly biased features, allowing it to retain more instances for downstream tasks. The feature weights are derived from the feature importance of a domain classifier trained to differentiate between the representative and non-representative datasets. We validated FW-MRS using eight tabular datasets, each of which we artificially biased. Biased features can be important for downstream tasks, and focusing less on them could lead to a decline in generalization. For this reason, we assessed the generalization performance of FW-MRS on downstream tasks and found no statistically significant differences. Additionally, FW-MRS was applied to a real-world dataset from the social sciences. The source code is available at \url{https://github.com/kramerlab/FeatureWeightDebiasing}.
\end{abstract}
\begin{document}

\flushbottom
\maketitle
\thispagestyle{empty}

\section{Introduction}
An ongoing challenge in the social sciences is that a sample may not accurately represent the broader population. When research relies on biased samples, it can lead to invalid conclusions and mistaken inferences about social processes \cite{winshipModelsSampleSelection1992}. An example of a biased study is one conducted in a specific city, despite the goal of gathering information about the entire country. If the study's bias is not addressed, researchers may misrepresent the population and draw inaccurate conclusions \cite{westHowBigProblem2016}.

Bias can occur at any stage of a research project, including study design, data collection, or data analysis \cite{smithSelectionMechanismsTheir2020,infante-rivardReflectionModernMethods2018}. It is not a dichotomous value but rather has varying intensities, and it is essential to estimate the degree of bias to interpret findings accurately, because using adequate debiasing  methods leads to more consistent findings and saves time and resources by avoiding repeated studies \cite{keebleChoosingMethodReduce2015}. The most effective way to minimize bias is to plan the study rigorously. However, even the best planning cannot always prevent it. When acquiring further data is not feasible, using algorithms to reduce  bias becomes the preferred option. While most times algorithms cannot completely remove bias, they can minimize it sufficiently to make the results usable.

\begin{figure}[ht]
    \centering
    \includegraphics[width=1\linewidth]{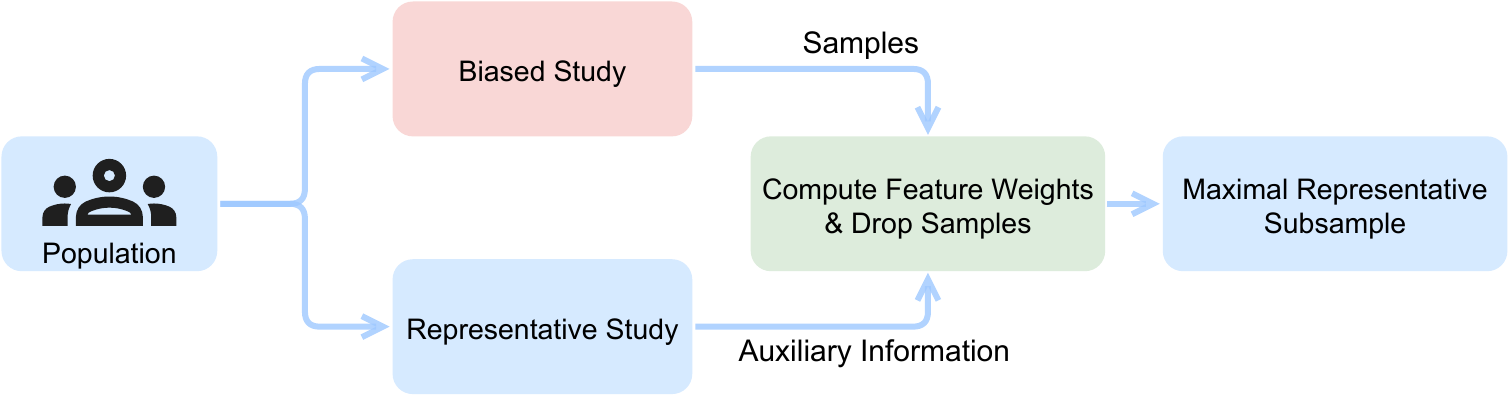}
    \caption{Schema for feature-weighted maximum representative sampling (FW-MRS). Two surveys stem from the same population: one is biased and includes the variable under investigation, while the other is representative but does not include it. FW-MRS mitigates bias by comparing the distributions of the biased and representative datasets using a classifier that leverages auxiliary information from the representative dataset to remove samples from the biased dataset. The algorithm returns a representative subset and feature weights that align the non-representative study to the distribution of the representative one.}
    \label{fig:fw-mrs_schema}
\end{figure}

A challenge for debiasing methods is the uneven distribution of bias across features: some are highly biased, whereas others are nearly unbiased. This is a problem because applying sample weights affects all features of a sample, not just the biased ones. Substantial changes to correct the distribution of biased features introduce bias into already representative features. In this work, we develop a method that tackles this challenge by incorporating feature weights into an algorithm that mitigates bias with sample weights. Instead of completely removing highly biased features, which could lead to the loss of valuable information, we propose a ``soft'' feature selection approach that utilizes feature weights. Our solution is based on \textit{Maximum representative subsampling} (MRS) \cite{hauptmannDiscriminativeMachineLearning2023}, which reduces bias by indirectly modelling the distributions of non-representative and representative samples using discriminative machine learning techniques. MRS removes elements from the biased sample with high probability of being non-representative and, in this way, iteratively aligns the distributions. MRS reduces the bias by generating a representative subsample of a dataset. It uses uniform weights that avoid assigning high values to samples that would otherwise dominate the analysis. \textit{Feature-weighted MRS} (FW-MRS) combines MRS with feature weights to reduce the influence of highly biased features, while still addressing the bias in less biased features. This strategy helps mitigate the effects of highly biased features, enabling us to retain more samples. The workflow of FW-MRS is illustrated in \Cref{fig:fw-mrs_schema}.

Our paper is organized as follows: First, we summarize  related work and provide an overview of MRS. Next, we introduce and explain the FW-MRS framework, presenting two variations based on different classification algorithms. In our experiments, we investigate how the temperature hyperparameter affects the number of dropped samples. We present findings from eight publicly available datasets in which we introduced artificial bias and performed a downstream classification. We validate the impact of the temperature hyperparameter on both the number of dropped samples and the downstream task's performance across these eight datasets. Subsequently, we apply FW-MRS to a real-world biased study and compare its results with MRS, focusing on the number of dropped samples and MMD after completion. Finally, we provide a summary and discussion of our results.
\section{Related Work}
Bias reduction algorithms exist in a great variety, and most of them rely on different types of additional information, such as class probabilities or computing sampling distributions to make corrections \cite{huangSupportVectorMachine2020,dudikCorrectingSampleSelection2005}. As distributions are rarely known and distribution estimation in high-dimensional spaces is complex, recent methods match distributions between the training and validation sets without performing density estimation. 

One approach is to compute a set of sample weights that transform the dataset's distribution and are used in the downstream task. \textit{Kernel Mean Matching} (KMM) \cite{huangCorrectingSampleSelection2006} is a non-parametric method that directly produces sampling weights without distribution estimation by reweighting the samples so that the means of the training and test samples in a reproducing kernel Hilbert space are close.  Another algorithm is \textit{propensity score adjustment} (PSA) \cite{rosenbaumCentralRolePropensity2006}, which trains a classifier to estimate the participation propensities of individuals in a non-representative dataset. The most commonly applied method in PSA is logistic regression. However, recent research has shown that different machine learning methods can be advantageous in estimating the propensity score \cite{ruedaEnhancingEstimationMethods2023}. 

A similar task to debiasing is domain adaptation, which addresses domain shifts. One established approach to domain adaptation with complex samples is to map the samples into a potentially lower-dimensional space where instances contain no domain information. Neural Networks are often used for that purpose, especially for images or text, but they are rarely used for tabular data. A well-known example of this is the \textit{Domain-adversarial Neural Network} (DANN) \cite{ganinDomainAdversarialTrainingNeural2016c}. It is a multi-target model that first embeds the data and splits into two classification branches: one for the domain and the other for the target. The gradient of the domain classification branch is inverted in the feature extractor to encourage training embeddings without domain information while preserving information essential for the target classification.

Not all features are equally representative of the hidden patterns, especially in real-world scenarios. To account for the differing amounts of information each feature provides, feature weights can be applied before model training. The primary advantage of feature weights lies in their ability to adjust the influence of each feature based on its estimated relevance to the predicted output. Feature weighting methods focus on evaluating a feature's importance in extracting the underlying pattern \cite{nino-adanFeatureWeightingMethods2021}.

Feature weights can be applied to samples by multiplication, creating an adjusted dataset that serves as input to a classifier \cite{nino-adanFeatureWeightingMethods2021}. Alternatively, scaling-invariant algorithms, such as \textit{random forests} (RF), can directly incorporate feature weights without modifying the dataset itself. These weights influence the algorithm's internal mechanisms. In feature-weighted random forests, the feature weights are integrated into the node-splitting process. In this case, features are selected based on relative weights rather than uniformly, allowing the model to prioritize the most informative features. This adjustment enhances the model’s predictive performance, particularly in high-dimensional data or weak-signal scenarios, such as gene expression data, and when incorporating prior knowledge \cite{amaratungaEnrichedRandomForests2008}.

This work presents FW-MRS, an extension of MRS that incorporates feature weights alongside sample weights to improve the alignment between two distributions. While existing weighting methods primarily focus on sample weighting, they overlook the influence of individual features on the domain shift. By integrating feature weights into the sample reweighting process, FW-MRS improves domain alignment while maintaining predictive performance.
\section{Feature-weighted Maximum Representative Subsampling}

The proposed method builds on MRS, which we  briefly summarize here (see again \Cref{fig:fw-mrs_schema}). MRS addresses dataset bias by removing biased samples to create a representative subsample, leveraging the distributional information of a representative dataset. The representative dataset must originate from the same population and share as many variables as possible with the biased dataset \cite{hauptmannDiscriminativeMachineLearning2023}.

MRS is based on \textit{positive-unlabeled} (PU) learning, a semi-supervised binary classification framework designed to work with only positive and unlabeled data \cite{bekkerLearningPositiveUnlabeled2020}. It requires two datasets: a positive set $P$ and a mixed set $U$ containing both positive and negative instances. The classifier is trained by treating all samples in $U$ as negative, even though some are positive, and is used to predict labels for an unlabeled test set $T$. MRS adapts PU learning by treating the representative dataset $R$ as $P$, treating the subset of the biased dataset $N$ as $U$, and treating the remaining portion as the test set $T$. A classifier is trained to distinguish $R$ from $N$, without explicitly modelling their distributions. Samples in $T$ that the classifier identifies as most likely non-representative are iteratively removed by setting their sample weights to zero. This process continues until the current classifier can no longer differentiate between the datasets. The final sample weights transform $N$ to a subset with approximately the same distribution as $R$ \cite{hauptmannDiscriminativeMachineLearning2023}. 

With this overview in mind, we now introduce the FW-MRS framework. First, FW-MRS computes feature weights from the feature importances computed by a domain classifier. The domain classifier is trained to distinguish representative from non-representative samples. Features that strongly differentiate between the two classes, non-representative ($N$) and representative ($R$), are considered highly biased and assigned lower weights. In contrast, features with low importance are considered less biased and receive higher weights. This transformation is accomplished through the softmin function with a temperature parameter. This approach helps the algorithm prioritize less biased features, thereby reducing the number of excluded samples.

We developed two FW-MRS variants, each using a different model and method to calculate feature importances. The first variant, FW-MRS$_{RF}$, employs a \textit{random forest} (RF) and calculates feature importances via SHAP values using TreeShap \cite{lundbergLocalExplanationsGlobal2020b} with interventional feature perturbation \cite{janzingFeatureRelevanceQuantification2020}. The second, FW-MRS$_{SVM}$, uses a linear \textit{Support Vector Machine} (SVM) with feature importances derived from Linear SHAP \cite{lundbergUnifiedApproachInterpreting2017b}. While the second approach is computationally less demanding, it detects only linear bias, making it suitable for scenarios where bias is expected to be linear or computational efficiency is a priority. The pseudo-code is provided in \Cref{alg:fw-mrs}. 

\begin{algorithm}[ht]
\KwIn{\\\Indp\Indp $N \in  \mathbb{R}^{m \times n}:$ non-representative set \\
$R \in \mathbb{R}^{r \times n}:$ representative set\\
$d:$ elements dropped per iteration \\
$k:$ cross-validation splits \\
$t:$ temperature \\
}
\KwResult{\\\Indp\Indp$w_s:$ \text{sample weights}\\
    $w_f:$ feature weights
}
$w = \{1\}^m $ \\
$I \leftarrow compute\_feature\_importances(classifier, N, R)$ \\

$w_f \leftarrow temperatured\_softmin(I,t)$ \\
\While{$R$ differs from $N$}{
    $auroc \leftarrow 0 $ \\
    $p \leftarrow [0]^m$ \\
    \For{$k$-fold splits}{
        $N_{train}, N_{test} \leftarrow N[split]$ \\
        $R_{train}, R_{test} \leftarrow R[split]$ \\
        classifier $\leftarrow$ train\_pu\_classifier(positive class = $N_{train}$, negative class = $R_{train}, w_s$, $w_f$) \\
        $p[test\_split] \leftarrow predict\_probabilities(classifier, N_{test})$ \\
        $auroc_{split} \leftarrow compute\_auroc(classifier, N_{test}, R_{test})$ \\
        $auroc \leftarrow auroc + auroc_{split}/k$ \\
    }
    \uIf{$auroc \leq 0.5$}{break} 
    \For{$i=0$ \KwTo $d$}{
        $drop\_index \leftarrow \underset{1 \leq i \leq n}{arg\,max}(p_i)$ \\
        $w_s[drop\_index] \leftarrow 0$ \\
        $p \leftarrow $removeElement$(p, {drop\_index}) $ \\
    }
}
\Return $w_s$, $w_f$
\caption{Pseudocode for Feature-Weighted Maximal Representative Subsampling.}
\label{alg:fw-mrs}
\end{algorithm}

The input to FW-MRS consists of the non-representative dataset ($N$), the representative dataset ($R$), the number of cross-validation splits ($k$), and the number of dropped samples per iteration ($d$). Like MRS, FW-MRS outputs sample weights, but additionally returns feature weights. The algorithm begins by initializing all sample weights uniformly (Line 1). A domain classifier is then trained to distinguish between samples from $N$ and $R$, and feature importances are computed using the chosen variant (Line 2). These importances are converted into feature weights using the softmin function:

\begin{equation}
    Softmin(I_i, t) = \frac{e^{\frac{-I_i}{t}}}{\sum_{j} e^{\frac{-I_j}{t}}},
\end{equation}

where $I_i$ denotes the feature importance of feature $i$, and $t$ is the temperature hyperparameter (Line 3). The softmin transformation ensures that more important, i.e., more biased, features receive lower weights. Lower values of $t$ produce a more peaked distribution, emphasising differences in importance, while higher values yield a more uniform weighting. In our experiments, $t$ is treated as a hyperparameter and optimized for the downstream task.

In the next step, a new domain classifier is trained, now incorporating both sample and feature weights. The implementation of this step differs by variant: In FW-MRS$_{RF}$, a feature- and sample-weighted random forest is trained. In FW-MRS$_{SVM}$, feature weights are applied by scaling the input features of all instances, modifying their influence during SVM training. Both variants employ $k$-fold cross-validation in conjunction with PU learning. For each fold, the classifier's outputs on the hold-out sets—either probabilities (RF) or decision function values (SVM)—are stored for later use (Lines 7–11). Additionally, the AUROC is computed for each fold to later assess the stopping criterion via the mean AUROC (Lines 12–13).

To monitor the stopping criterion, the algorithm tracks the mean AUROC across folds (Lines 15-16). If the AUROC falls below $0.5$, indicating the classifier is not better than random guessing, FW-MRS terminates, assuming sufficient distribution alignment. Otherwise, the $d$ samples most confidently identified as non-representative are removed by setting their weights to zero (Lines 17-21). The process then repeats with updated sample weights. Once the stopping criterion is fulfilled, FW-MRS returns both the sample and feature weights (Line 23). These weights can be used to reweight the non-representative dataset for the downstream task.
\section{Experimental Results}
This section presents our experiments comparing FW-MRS with MRS and other debiasing methods. We validated the approaches on eight tabular datasets, each featuring a binary classification task comprising at least several hundred samples. The datasets span the social and life sciences, where debiasing is most commonly needed. Specifically, we used: Folktables Income and Folktables Employment \cite{dingRetiringAdultNew2021} (for brevity called Income and Employment), Human Analytic, and the  Loan dataset from Kaggle. Breast Cancer (Wisconsin), German Credit \cite{hofmannStatlogGermanCredit1994}, Diabetes \cite{kahnDiabetes00} and Bank Marketing \cite{s.moroBankMarketing2014} from the UCI Repository \cite{misc_breast_cancer_wisconsin_diagnostic_17}. The selected datasets are publicly available, contain demographic information, and are presented in a tabular format with a binary classification task. Details about the dataset characteristics can be found in the Supplementary Table 1.

\subsection{Experimental Setup}
Our evaluation procedure consists of the following steps: First, each dataset is randomly split using 5-fold cross-validation. The training set is then divided into two equal parts: one is artificially biased to serve as the non-representative dataset $N$. At the same time, the other remains unchanged and acts as the representative dataset $R$. The test set is kept separate and unaltered throughout. To ensure robustness, the entire cross-validation process is repeated 10 times. During training, the class labels are only available for $N$; for $R$, the assumption is that no class labels are available.

Artificial bias is introduced by undersampling the positive class in $N$, retaining only 10\% of its original positive samples. This simulates real-world scenarios, such as underdiagnosed medical conditions misrepresenting the actual prevalence, or electoral studies where marginalized groups are undercounted, skewing turnout statistics. To keep data sizes manageable, we randomly sampled 6,000 instances from the larger datasets (Diabetes, Employment Income, Bank Marketing, and HR Analytics) before conducting cross-validation. For these datasets, five samples were removed per iteration; for the smaller datasets, one sample was removed per iteration.

For the downstream classification task, we trained a feature-weighted random forest with 500 trees on $N$ using the sample and feature weights returned by the debiasing method. The downstream classifier was then used to evaluate performance on the test set. The hyperparameters of the downstream classifier and the debiasing were optimized on $N$, using the AUROC of the downstream task as the model selection criterion. FW-MRS$_{RF}$ employs a random forest with 200 decision trees and was optimized over the temperature [0.001, 0.0025, 0.005, 0.01, 0.025, 0.05, 0.1, 0.25, 0.5] and the minimum weight fraction per leaf [0.025, 0.01, 0.001, 0.0]. FW-MRS$_{SVM}$ uses a linear SVM with $l_2$ regularization and C was optimized over $[1e-2, ..., 1e2]$. Both variants employ PU learning with 5-fold cross-validation.

To confirm that FW-MRS not only makes data sets indistinguishable but also increases the similarity of the distributions, the \textit{maximum mean discrepancy} (MMD), which measures the distance between two distributions, is used. MMD can be computed as the norm of the difference between the feature means of the distributions in the reproducing kernel Hilbert space \cite{rkhs}. 

Given two sets of samples $X= \{x_i\}_{i=1}^n$ and $Y=\{y_i\}_{i=1}^m$, one can compute the empirical estimate of the MMD with sample weights in the following way:

\begin{equation}
    \begin{aligned}
         \text{MMD}(X, Y, w_X, w_Y, w_f) = \left[\sum_{i,j=1}^{m} w_{X}^{(i)}w_{X}^{(j)} K(x_i, x_j, w_f) - 2 * \sum_{i,j=1}^{m,n} w_{X}^{(i)}w_{Y}^{(j)} K(x_i, y_j, w_f) +  \sum_{i,j=1}^{n} w_{Y}^{(i)}w_{Y}^{(j)} K(y_i, y_j, w_f)\right]^{\frac{1}{2}}
     \end{aligned}
\end{equation}

where $w_X$ and $w_Y$ are the sample weights, $w_f$ are the feature weights, and $k$ is the weighted radial basis function kernel:

\begin{equation}
         k(x, y, w_f) = \exp(- \frac{\sum_j w_f^{(j)} (x_j-y_j)^2}{2\sigma^2}).
\end{equation}

We determine $\sigma$ heuristically, setting it to the mean distance between the samples in the aggregated sample, which contains all samples from both data sets \cite{mmd}.

We compared FW-MRS to uniform weighting, the original MRS, and the two most important and widely used sample-weight-based bias-reduction methods. The first baseline is KMM, with $\sigma$ heuristically selected by calculating the mean pairwise distance between samples in the combined dataset using the radial basis function kernel. The second is PSA with logistic regression, where sample weights are computed as inverse propensity score: $(1 - \pi) / \pi$ \cite{schonlauOptionsConductingWeb2017}. Logistic regression employed $l_2$-regularization, and the regularization parameter $C$ was optimized over the same range as above.

Neural network-based methods were not included in the comparison, as we focused on tabular data (rather than images or text) and did not want to add a layer of complexity due to the expense of hyperparameter optimization. Furthermore, we examine feature and instance weights, which can be inspected and visualized, rather than NN-based embeddings or latent representations.

\subsection{Temperature Comparison}
\begin{figure}[ht]
    \centering
    \includegraphics[width=0.7\linewidth]{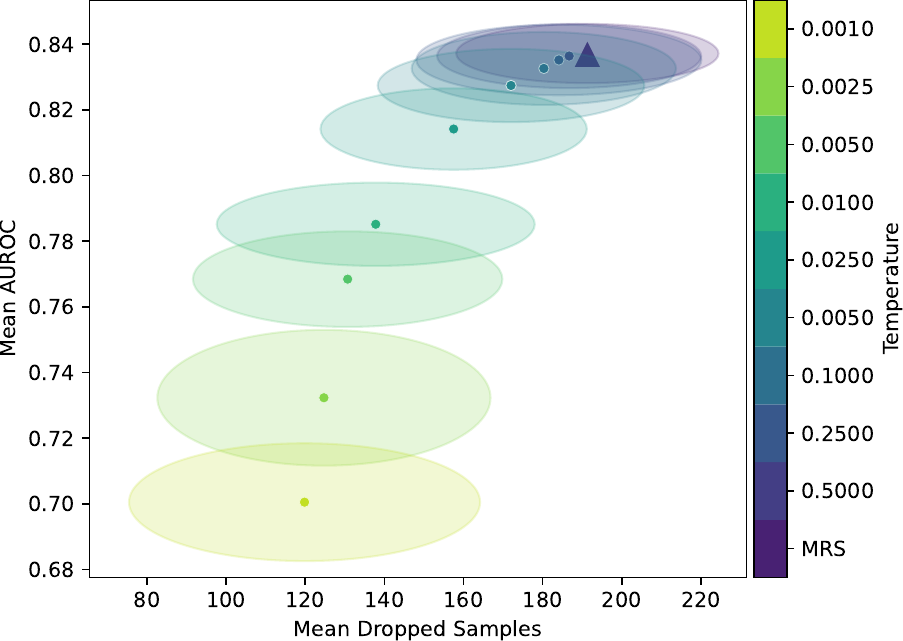}
    \caption{Validation AUROC vs. number of dropped samples: All hyperparameters were fixed except for the temperature, which was varied. Each point denotes the mean AUROC across runs, with ellipses indicating the standard deviation. Circles correspond to FW-MRS$_{RF}$ and the triangle represents MRS.}
    \label{fig:temperature_comparison_fixed}
\end{figure}

In this section, we visualize how the temperature parameter $t$ influences the number of dropped samples and the downstream task performance. In the first experiment, we evaluated FW-MRS$_{RF}$ by varying the temperature while keeping all other hyperparameters fixed. For each temperature setting, we recorded the number of samples dropped and computed the AUROC on $R$ as a validation metric. We performed the same procedure for MRS to work as a baseline. The results are shown in \Cref{fig:temperature_comparison_fixed}, where the mean number of dropped samples is plotted against the mean AUROC. Each point represents the average across runs, with ellipses indicating standard deviations. FW-MRS$_{RF}$ results are shown as circles, and MRS as triangles.

The diagram reveals a non-linear correlation between the number of dropped samples and the AUROC computed with different temperatures. Initially, as the temperature decreases, both the number of dropped samples and the AUROC decline gradually. This early phase reflects the reduced influence of biased features: although they are weighted less, they are still used sufficiently that downstream task performance remains largely unaffected. As the temperature is further reduced, feature weighting becomes more extreme, and the AUROC declines more sharply, indicating that informative but biased features are now underutilized. Meanwhile, the number of dropped samples continues to fall more rapidly. Eventually, the process reaches a point at which most biased features are used infrequently, resulting in only marginal reductions in dropped samples. At this stage, the performance loss outweighs the benefit of retaining more samples. This trade-off is also reflected in the increased variance in AUROC and the number of dropped samples at lower temperatures.

\begin{figure}[ht]
    \centering
    \includegraphics[width=0.8\linewidth]{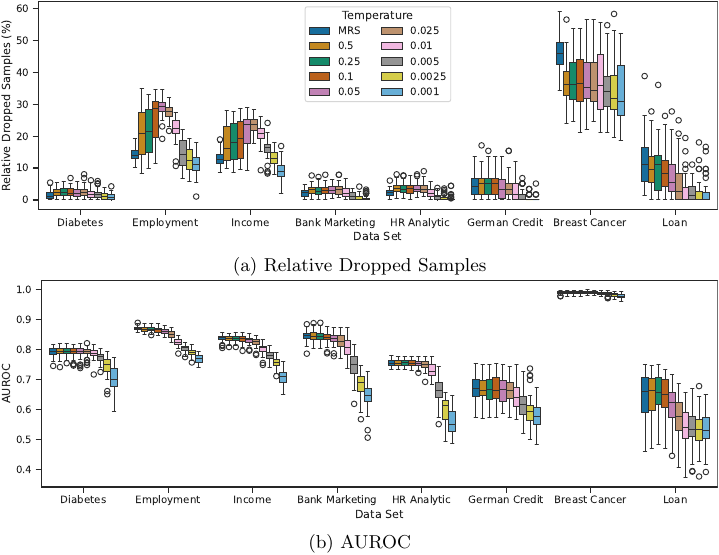}
    \caption{Relative dropped samples and AUROC for 50 iterations with 10 times repeated 5-fold cross-validation for FW-MRS$_{RF}$. Subfigure a) shows the relative dropped samples, and b) the corresponding AUROC. Hyperparameters were optimized on $N$ for the downstream task, and the effect of different temperature settings is compared.}
    \label{fig:dropped_samples_per_temperature}
\end{figure}

In the next experiment, we analysed the effect of temperature on the number of dropped samples and downstream task performance using box plots of the 50 values recorded across iterations for each temperature setting and MRS (\Cref{fig:dropped_samples_per_temperature}). In many cases, MRS consistently discards the most samples, especially at lower temperatures. In general, incorporating feature weights results in fewer dropped samples, and as the temperature decreases, the number of dropped samples generally declines, though a few exceptions exist. One notable exception with a high rate of dropped samples is Breast Cancer, where all methods tend to drop many samples. This is likely because the dataset has few features, all of which are ordinal, making it challenging to perform debiasing efficiently. 

The Breast Cancer and Loan datasets illustrate the impact of feature weighting particularly well. Due to its small size and entirely ordinal features, MRS must discard a substantial portion of the data to align the distributions. In contrast, incorporating feature weights significantly reduces the number of dropped samples in these cases.

We also examined the influence of temperature on AUROC. The trend is even more pronounced: lower temperatures result in a noticeable decline in AUROC, which drops sharply once the temperature falls below a specific threshold. This decline is likely due to the nature of the simulated bias, in which the biased features are also predictive of the downstream task. At low temperatures, some of these informative features are given minimal weight, causing them to be nearly ignored by the random forest model, effectively removing them from the model. This behaviour resembles feature elimination and degrades performance.

\subsection{Downstream Task}
After investigating how feature weights influence the number of dropped samples and AUROC in general, we assess their impact on downstream classification  performance and the number of dropped samples when optimizing the temperature and hyperparameters for the downstream task. As shown in the previous experiment, a decline in AUROC is expected. However, if the difference is not statistically significant, FW-MRS offers a favourable alternative. The first experiment measures the impact of debiasing on the performance of the downstream task.

\begin{figure}[ht]
    \centering
    \includegraphics[width=0.8\linewidth]{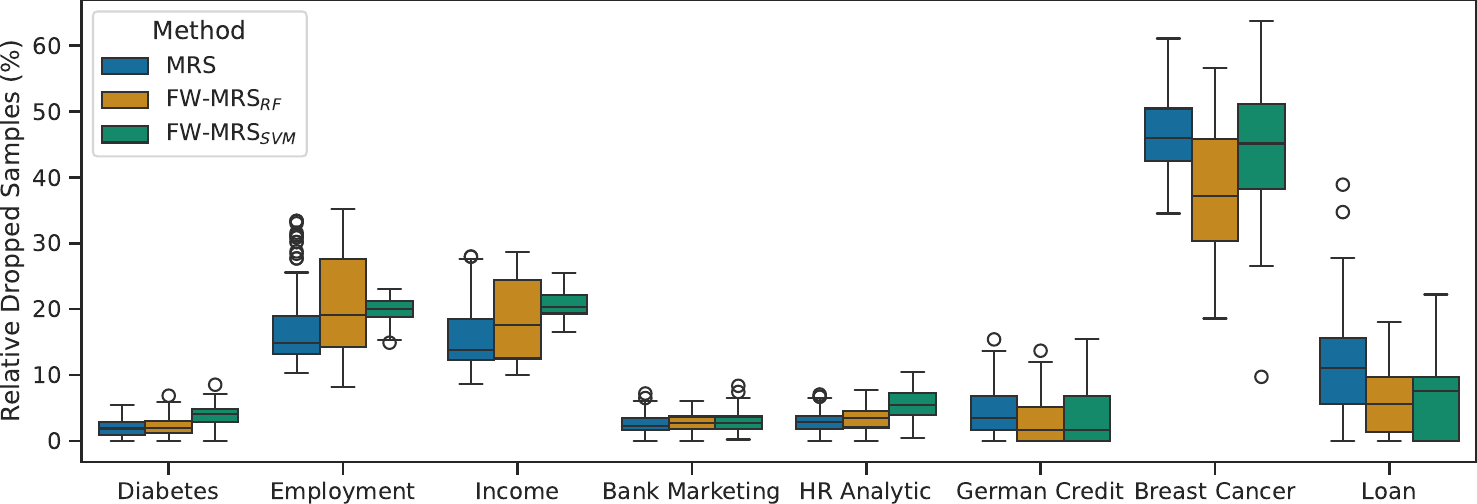}
    \caption{Relative dropped samples over 50 iterations with 10 times repeated 5-fold cross-validation.}
    \label{fig:downstream_dropped_elements}
\end{figure}

First, we visually examined the relative number of dropped samples during the downstream task, treating the temperature as a hyperparameter. \Cref{fig:downstream_dropped_elements} displays the box plots along with the corresponding relative number of dropped samples for each method. It is visible that FW-MRS$_{RF}$ retained more samples than MRS in five of eight datasets, while FW-MRS$_{SVM}$ did so in four. Overall, both FW-MRS variants tend to keep more samples, particularly on smaller datasets.

Second, the mean AUROC and standard deviation are reported in \Cref{tab:downstream_metrics}. Based on the mean rank, MRS achieved the second-highest average AUROC, following uniform weighting. PSA and KMM showed the largest decrease in predictive performance, likely due to assigning extreme sample weights that significantly altered their influence during training. FW-MRS$_{SVM}$ resulted in a greater reduction in AUROC than FW-MRS$_{RF}$, whose performance remained comparable to that of MRS. These results indicate that although feature weighting can reduce downstream performance, the proposed methods perform competitively. Overall, the mean ranking should be interpreted cautiously, as the differences in mean AUROC across methods are minor, and the standard deviation intervals largely overlap. 

We hypothesize that the reduced generalization is due to the bias introduced by undersampling the positive class. This leads to a situation where features correlated with the downstream task are especially biased. Since the bias is related to the target variable, essential features may be systematically undervalued, explaining the observed decline in predictive performance. In this setting, using uniform feature weights yielded the best downstream results, likely because both sample removal and feature downweighting entail information loss.

\begin{table}[ht]
    \centering
    \caption{Downstream task AUROC over 50 iterations with 10 times repeated 5-fold cross-validation. The numbers are the means and standard deviations. The best values are written in bold, and the second best is  underlined. No statistically significant differences between FW-MRS and MRS could be detected. The significance was tested with a corrected \textit{t}-test and the Benjamini-Hochberg procedure. For Unbiased metrics, the downstream classifier was trained on an unbiased dataset.}
    \begin{adjustbox}{max width=\textwidth}
    \begin{tabular}{l cc cc cc|c}
        \toprule
         Dataset & Uniform & KMM & PSA & MRS & FW-MRS$_{RF}$ & FW-MRS$_{SVM}$ & Unbiased \\
         \midrule
            Diabetes       & $\mathbf{0.791\pm0.02}$ & $0.781\pm0.02$ & $0.787\pm0.02$ & $\underline{0.789\pm0.02}$ & $0.784\pm0.02$ & $0.782\pm0.03$ & $0.811\pm0.01$ \\
            Employment     & $\mathbf{0.871\pm0.01}$ & $0.857\pm0.01$ & $0.867\pm0.01$ & $\underline{0.870\pm0.01}$ & $0.868\pm0.01$ & $0.864\pm0.01$ & $0.888\pm0.01$ \\
            Income         & $\mathbf{0.838\pm0.01}$ & $0.820\pm0.01$ & $0.831\pm0.01$ & $\underline{0.837\pm0.01}$ & $0.836\pm0.01$ & $0.833\pm0.01$ & $0.866\pm0.01$ \\
            Bank Marketing & $\mathbf{0.847\pm0.02}$ & $0.832\pm0.03$ & $0.839\pm0.03$ & $\underline{0.845\pm0.02}$ & $0.843\pm0.02$ & $0.840\pm0.02$ & $0.906\pm0.01$ \\
            HR Analytic    & $\mathbf{0.753\pm0.02}$ & $0.749\pm0.02$ & $0.750\pm0.02$ & $\underline{0.751\pm0.02}$ & $0.750\pm0.02$ & $\underline{0.751\pm0.02}$ & $0.764\pm0.02$ \\
            German Credit  & $\underline{0.667\pm0.05}$ & $0.649\pm0.05$ & $0.659\pm0.06$ & $\mathbf{0.672\pm0.05}$ & $0.642\pm0.06$ & $0.639\pm0.07$ & $0.764\pm0.03$ \\
            Breast Cancer  & $\underline{0.988\pm0.01}$ & $\mathbf{0.989\pm0.01}$ & $\underline{0.988\pm0.01}$ & $\mathbf{0.989\pm0.01}$ & $0.981\pm0.01$ & $0.977\pm0.01$ & $0.992\pm0.00$ \\
            Loan           & $\mathbf{0.658\pm0.08}$ & $0.610\pm0.10$ & $0.628\pm0.10$ & $\underline{0.645\pm0.09}$ & $0.612\pm0.08$ & $0.586\pm0.10$ & $0.753\pm0.05$ \\
            \midrule
            Rank & \textbf{1.44} & 5.06 & 3.88  & \underline{1.88} & 3.94 & 4.81 & $\emptyset$ \\
        \bottomrule
    \end{tabular}
    \end{adjustbox}
    \label{tab:downstream_metrics}
\end{table}

To assess whether the differences between MRS and FW-MRS are statistically significant, we performed a corrected repeated $k$-fold cross-validation \textit{t}-test \cite{bouckaertEvaluatingReplicabilitySignificance2004}. This test adjusts the variance using a correction factor to account for the inflated Type I error rate introduced by overlapping training sets. Given the multiple comparisons, we adjusted the \textit{p}-values using the Benjamini–Hochberg procedure, which reduces the number of false positives by setting \textit{p}-values to consider only the most reliable results as significant. An $\alpha < 0.05$ was regarded as statistically significant. Although both FW-MRS variants slightly decreased the mean AUROC, none of the differences were statistically significant. This indicates that incorporating feature weights has a negligible impact on downstream performance in most cases. 

Both experiments together highlight that while fewer samples could always be dropped, the optimization procedure inherently prioritizes performance over sample preservation. If preserving samples is the primary objective, the optimization process must be adjusted accordingly. Alternatively, it may be a combination of both. Reducing the temperature not only retains more samples but also reduces the MMD between the debiased dataset $N$ and the representative test set $T$, as shown in the Supplementary Table 3.

\subsection{Real-world Dataset}

In this experiment, we apply FW-MRS$_{RF}$ to a real-world case and analyse the influence of the temperature on the number of dropped samples and MMD. The real-world dataset is part of the \textit{Gutenberg Brain Study} (GBS), a population-based study designed to examine how resilience influences voting behaviour. The study was carried out in accordance with relevant guidelines and regulations. The study protocol was approved by the ethics committee at the Rhineland-Palatinate state chamber of physicians (No 837.085.13, 8770-F) and participants signed a consent form to obtain written consent. 

However, a significant limitation of this study is that it was conducted in a university city, which does not accurately reflect the country's overall demographic diversity. Additionally, the study protocol may be subject to self-selection bias, as individuals with a particular interest in the topic are more likely to participate in the interview.

To mitigate the bias, we used a representative dataset from the Allensbach Institute for Public Opinion Research (Allensbach). The individuals were selected as they met criteria of the quota sample based on the German official statistics. Thus the data can be generalized to the German population with normal three percent of statistical uncertainty within representative surveys. In order to ensure informed consent, participants were informed about the objectives of the study, procedures of data storage and protection and their right to withdraw from the study at any point in time. They were informed that their participation is voluntary. Verbal consent was obtained to ensure anonymity. The interviewers had a standardized questionnaire and could answer further questions for example if there were uncertainties. The study was approved by the ethics committee at the Rhineland-Palatinate state chamber of physicians (No 837.209.14, 9448F) and the participation in the survey was voluntary.

Auxiliary information in the dataset was employed to minimize bias in the GBS using FW-MRS$_{RF}$. The debiasing was repeated 50 times, and the mean feature importance of the unweighted feature importances and the mean of the derived feature weights were calculated. Additionally, the mean number of dropped samples per temperature and the MMD between the weighted GBS and Allensbach were computed. The hyperparameters of FW-MRS$_{RF}$, except the temperature, were optimized to minimize the MMD between GBS and Allensbach. We also tracked the mean AUROC across runs and plotted it against the number of iterations in Supplemental Figure 1.

The feature weights provide a way to measure the degree of bias in each feature, and the mean feature weights with standard deviation are visualized in \Cref{fig:gbs_allensbach_feature_weights} to provide an overview (see the Supplemental Figure 2 for a visualization of the feature importances). The features are ordered from left to right in ascending order of feature importance, i.e., from least biased to most. As expected, employment status, educational attainment, and occupational group are the most biased features, given the city's higher educational level. For FW-MRS$_{RF}$, it starts with a nearly uniform distribution, similar to MRS. However, by lowering the temperature, the weights are shifted to the features with the lowest importance, until, at some point, all the weights are assigned to the feature of minimum importance. 

\begin{figure}[ht!]
    \centering
    \includegraphics[width=0.7\linewidth]{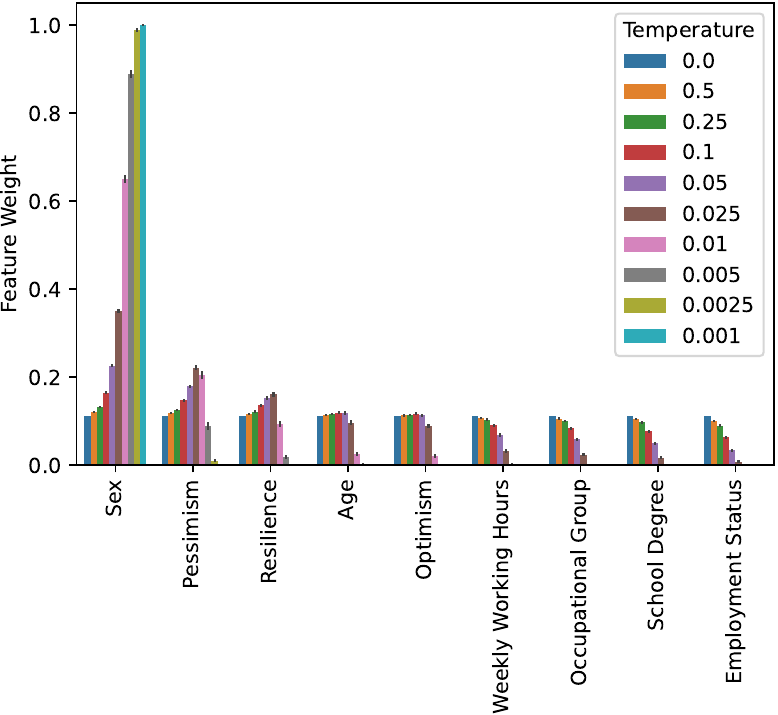}
    \caption{Feature weights used in FW-MRS$_{RF}$ debiasing of GBS with auxiliary information of Allensbach.}
    \label{fig:gbs_allensbach_feature_weights}
\end{figure}

\Cref{tab:gbs_stats} contains the mean MMD and number of dropped samples for FW-MRS$_{RF}$ with different temperatures and MRS. As the temperature decreases, more samples are retained, and the MMD between GBS and Allensbach decreases. However, choosing the lowest temperature is not advisable, as it assigns feature weights to mostly one feature (sex), which alone is not informative.

\begin{table}[ht!]
    \centering
    \begin{tabular}{l ccc}
        \toprule
         Temperature & Number of Dropped Samples & MMD \\
         \midrule
         MRS    & $501.1 \pm 4.9$ & $0.153 \pm 0.004$  \\
         0.5    & $501.0 \pm 3.6$ & $0.151 \pm 0.004$  \\
         0.25   & $497.5 \pm 3.6$ & $0.145 \pm 0.003$  \\
         0.1    & $493.2 \pm 4.2$ & $0.132 \pm 0.005$  \\
         0.05   & $478.7 \pm 5.1$ & $0.114 \pm 0.004$  \\
         0.025  & $444.5 \pm 4.9$ & $0.086 \pm 0.002$  \\
         0.01   & $305.8 \pm 12.7$ & $0.059 \pm 0.001$ \\
         0.005  & $235.8 \pm 15.3$ & $0.039 \pm 0.002$ \\
         0.0025 & $152.2 \pm 9.7$ & $0.034 \pm 0.002$ \\
         0.001  & $143.0 \pm 7.6$ & $0.032 \pm 0.001$ \\
         \bottomrule
    \end{tabular}
    \caption{Mean number of dropped samples and MMD of 50 runs debiasing of GBS with auxiliary information of Allensbach with FW-MRS$_{RF}$.}
    \label{tab:gbs_stats}
\end{table}

The ideal temperature selection varies depending on the specific situation. Researchers need to balance the trade-off between the number of dropped samples and the MMD, while accounting for the weights assigned to different features. This can be done either manually or using a compound metric for optimization. Selecting a temperature that is too low can result in the loss of too much information, especially if some features receive almost no weight. Conversely, selecting a temperature that is too high may result in the removal of too many samples, thereby adversely affecting statistical power.

To summarize our findings:
\begin{itemize}
    \item There is no evidence that the downstream performance of FW-MRS is statistically significantly different from that of MRS, the currently best method based on downstream AUROC (\Cref{tab:downstream_metrics} and the results of significance tests).
    \item FW-MRS retains more instances and thus reduces in some cases the variance component of the error compared to MRS (see the lower part of Supplementary Table 2).
    \item Considering the distribution alignment, FW-MRS improves on MRS (see Supplementary Table 3). While KMM and PSA achieve even stronger alignment, this comes at the cost of a greater decline in downstream performance (\Cref{tab:downstream_metrics}), suggesting a less favorable trade-off.
    \item Temperature selection is critical, and can be supported with visualizations similar to \Cref{fig:temperature_comparison_fixed} or determined via hyperparameter optimization.
\end{itemize}
\section{Discussion}
In this paper, we introduced the FW-MRS framework  and presented two specific implementations: one based on random forests (FW-MRS$_{RF}$) and the other on linear SVMs (FW-MRS$_{SVM}$). FW-MRS extends the original MRS approach by incorporating feature weights that control the influence of biased features during both downsampling and downstream classification.

By downweighting biased features, FW-MRS reduces the number of samples that must be discarded. FW-MRS$_{RF}$ has less impact on downstream task performance than FW-MRS$_{SVM}$. High temperatures were mainly used when optimizing hyperparameters for downstream classification to minimize their effect on downstream task quality. Still, if a fixed number of samples should be retained, e.g., to maintain statistical power, the optimization objective can be adjusted to prioritize the number of dropped samples or use a mixture of both.

FW-MRS is relevant in settings where aligning disparate data distributions is essential to train representative, generalizable models. For instance, in healthcare, datasets are collected mostly independently by institutions and may vary substantially. FW-MRS provides an approach to align data from diverse sources, creating more reliable predictive models. FW-MRS can be used to align the datasets, enabling their combination in the downstream task and potentially improving predictive performance.  When targets are available for both datasets, this information could be further integrated into the method to enhance alignment.

The FW-MRS framework is designed to be flexible. It can be customized to specific needs by substituting the classifier, redefining feature importance measures, or adapting the hyperparameter optimization strategy. The method's adaptability enables more representative and reliable inference from datasets affected by strong feature biases.

\FloatBarrier

\bibliography{mybibfile}

\FloatBarrier
\section*{Author Contributions}
Tony Hauptmann implemented methods and experiments and wrote the manuscript. Stefan Kramer supervised the study, developed the concept, and wrote and reviewed the manuscript.

\section*{Competing Interests}
All authors declare that they have no competing interests relevant to the research presented in this article. 

\section*{Data Availability}
GBS and Allensbach are private datasets. Human Resource Analytics (\url{https://www.kaggle.com/datasets/arashnic/hr-analytics-job-change-of-data-scientists}) and Loan (\url{https://www.kaggle.com/datasets/burak3ergun/loan-data-set}) are freely available on Kaggle. Folktables can be accessed through its Python package (\url{https://github.com/socialfoundations/folktables}). Breast Cancer (Wisconsin) (\url{https://archive.ics.uci.edu/dataset/15/breast+cancer+wisconsin+original}), German Credit (\url{https://archive.ics.uci.edu/dataset/144/statlog+german+credit+data}), Diabetes (\url{https://archive.ics.uci.edu/dataset/34/diabetes}), and Bank Marketing (\url{https://archive.ics.uci.edu/dataset/222/bank+marketing}) are available at the UCI Repository.

\section*{Code Availability}
The source code for preprocessing GBS and Allensbach (\url{https://github.com/laksannathan/maximal-representative-sampling}) and the methods and experiments  (\url{https://github.com/kramerlab/FeatureWeightDebiasing}) are available.

\section*{Funding}
The German Federal Ministry of Education and Research funded this work under grant number [16LW0242] as part of the DIASyM project.

\appendix

\section{Dataset Characteristics}

\begin{table}[h!]
    \caption{Dataset characteristics.}
    \centering
    \label{tab:dataset_characteristics}
    \begin{adjustbox}{max width=\columnwidth}
    \begin{tabular}{l cc cc}
    \toprule
         Name & \#Samples & \#Positive & \#Negative & \#Features \\
         \midrule
         Diabetes & 253680 & 35346 & 218334 & 21 \\
         Folktables Employment & 378817 & 172803 & 206014 & 99 \\
         Folktables Income & 195665 & 85189 & 110476 & 69 \\
         Bank Marketing & 45211 & 5289 & 39922 & 16 \\
         Human Research Analytic & 8955 & 1483 & 7472 & 33\\
         Allensbach & 1082 & $\emptyset$ & $\emptyset$ &  54 \\
         German Credit & 1000 & 300 & 700 & 20 \\
         Breast Cancer & 683 & 444 & 239 & 10 \\
         GBS & 579 & 550 & 29 & 52 \\
         Loan & 480 & 332 & 148 & 13 \\
         \bottomrule
    \end{tabular}  
    \end{adjustbox}
\end{table}

\section{Bias-Variance Decomposition}
Next, we analyze the bias-variance decomposition of the 0-1 loss. To perform the decomposition,  the test procedure was modified: First, each dataset was randomly split into two equal parts: one half referred to as the "world", with a size of $2m$, was used to draw training samples from, while the other half was used as a fixed test set. From the world, we sampled $m$ samples without replacement to form a training set. We introduced a bias within each sample by selecting a non-representative subset $N$, with the remaining instances forming the representative dataset $R$. This sampling approach allows  $\binom{2n}{n}$ distinct datasets, providing sufficient variety \cite{kohaviBiasVarianceDecomposition1996}. The sampling was repeated 50 times, but the test set was kept constant to facilitate the bias-variance decomposition (\Cref{tab:downstream_bias_variance_01}).

The bias and variance of the 0-1 loss were computed in the following: Given the main prediction $E[\hat{y}]$ as the mode of the predicted classes and $L(y, \hat{y})$ defines the 0-1 loss, the bias is computed as $L(y, E[\hat{y}])$ and the variance as $E[L(\hat{y}, E[\hat{y}])]$ with the expectation taken over training sets \cite{domingosUnifiedBiasVarianceDecomposition2000}.

\begin{table}[ht!]
    \centering
    \caption{Downstream task bias and variance for 0-1 loss over 50 iterations using repeated resampling. The numbers are the means, the best values are written in bold, and the second best is underlined. No value is underlined in rows where all methods perform equally.}
    \begin{adjustbox}{max width=\textwidth}
    \begin{tabular}{ll cc cc cc|c}
         \toprule
         Metric & Dataset  & Uniform & KMM & PSA & MRS & FW-MRS$_{RF}$ & FW-MRS$_{SVM}$ & Unbiased \\
         \midrule
         \multirow{8}{*}{Bias}
            & Diabetes       & \textbf{0.135} & \textbf{0.135} & \textbf{0.135} & \textbf{0.135} & \textbf{0.135} & \textbf{0.135} & 0.208 \\
            & Employment     & \textbf{0.212} & 0.262 & 0.213 & \underline{0.216} & 0.229 & 0.224 & 0.205 \\
            & Income         & 0.292 & \textbf{0.252} & 0.288 & \underline{0.278} & 0.291 & 0.286 & 0.208 \\
            & Bank Marketing & \textbf{0.116 }& \textbf{0.116} & \textbf{0.116} & \textbf{0.116} & \textbf{0.116} & \textbf{0.116} & 0.097 \\
            & HR Analytic    & \underline{}{0.165} & \underline{0.165} & \underline{}{0.165} & \textbf{0.155} & \underline{0.165} & \underline{0.165} & 0.165 \\
            & German Credit  & \textbf{0.299} & \textbf{0.299} & \textbf{0.299} & \textbf{0.299} & \textbf{0.299} & \textbf{0.299} & 0.257 \\
            & Breast Cancer  & 0.048 & \textbf{0.031} & \underline{0.035} & \underline{0.035} & 0.044 & 0.039 & 0.018 \\
            & Loan           & \textbf{0.694} & \textbf{0.694} & \textbf{0.694} & \textbf{0.694} & \textbf{0.694} & \textbf{0.694} & 0.206 \\
                \midrule
                Rank & & 3.88 & \underline{3.25} & 3.31 &  \textbf{2.81} & 4.13 & 3.63 & $\emptyset$ \\ 
         \midrule
         \multirow{8}{*}{Variance} 
                & Diabetes       & 0.076 & \textbf{0.053} & 0.094 & \underline{0.070} & 0.080 & 0.084 & 0.105 \\
                & Employment     & 0.157 & 0.200 & \textbf{0.143} & 0.160 & \underline{0.148} & 0.161 & 0.029 \\
                & Income         & \underline{0.134} & 0.256 & 0.217 & \textbf{0.133} & 0.143 & 0.141 & 0.052 \\
                & Bank Marketing & 0.014 & 0.034 & 0.027 & 0.014 & \underline{0.013} & \textbf{0.012}  & 0.025 \\
                & HR Analytic    & \underline{0.072} & 0.157 & 0.109 & 0.089 & 0.075 & \textbf{0.071}  & 0.046 \\
                & German Credit  & \textbf{0.014} & 0.020 & 0.025 & 0.020 & \underline{0.017} & 0.037  & 0.147 \\
                & Breast Cancer  & 0.104 & 0.077 & \underline{0.040} & \textbf{0.017} & 0.055 & 0.052 & 0.007 \\
                & Loan           & \underline{0.066} & 0.091 & \textbf{0.064} & \underline{0.066} & 0.092 & 0.082 & 0.034 \\
            \midrule
            Rank & &  \underline{2.88} & 4.81 & 3.75 & \textbf{2.69} & 3.38 & 3.50 & $\emptyset$ \\ 
         \bottomrule
    \end{tabular}
    \end{adjustbox}
    \label{tab:downstream_bias_variance_01}
\end{table}

MRS achieves the greatest reduction in both bias and variance, although all methods exhibit the same bias across four datasets. PSA and KMM tend to increase variance due to the variability introduced by potentially widely varying sample weights. In contrast, (FW)-MRS assigns uniform weights to all remaining samples, mitigating this effect. However, both FW-MRS variants exhibit increased bias, as the downstream classifier places less weight on informative, biased features. FW-MRS decreases the variance in some cases and increases it in others.

\newpage
\section{Distribution Alignment}
In this experiment, the feature- and sample-weighted MMD between the debiased non-representative dataset $N$ and the representative test set $T$ is computed and compared across different methods. As expected, uniform weighting yielded the worst results (\Cref{tab:distribution_adaptation_metric}). The remaining methods demonstrate varying degrees of improved distribution alignment. MRS achieved moderate alignment, whereas both FW-MRS variants, especially FW-MRS$_{SVM}$, further reduce the MMD, indicating that feature weights improve the distribution alignment. PSA and KMM achieved the lowest MMD values, especially KMM, which explicitly optimises MMD. However, this came at the cost of a greater decline in downstream task performance for KMM compared to MRS and FW-MRS.

\begin{table}[h!]
    \centering
    \caption{MMD over 50 iterations with 10 times repeated 5-fold cross-validation. The numbers are the means and standard deviations. The best values are written in bold, and the second best is underlined.}
    \begin{adjustbox}{max width=\textwidth}
    \begin{tabular}{l cc cc cc |c}
        \toprule
         Dataset & Uniform & KMM & PSA  & MRS & FW-MRS$_{RF}$ & FW-MRS$_{SVM}$ & Unbiased \\
         \midrule
    Diabetes       & $0.0358\pm0.006$ & $\mathbf{0.0217\pm0.003}$ & $\underline{0.0220\pm0.003}$ & $0.0325\pm0.005$ & $0.0319\pm0.005$ & $0.0308\pm0.006$ & $0.0220\pm0.003$ \\
    Employment     & $0.0681\pm0.005$ & $\mathbf{0.0219\pm0.002}$ & $\underline{0.0240\pm0.002}$ & $0.0472\pm0.013$ & $0.0465\pm0.014$ & $0.0329\pm0.005$ & $0.0224\pm0.002$ \\
    Income         & $0.0640\pm0.005$ & $\mathbf{0.0233\pm0.003}$ & $\underline{0.0262\pm0.002}$ & $0.0501\pm0.011$ & $0.0514\pm0.011$ & $0.0342\pm0.003$ & $0.0230\pm0.003$ \\
    Bank Marketing & $0.0282\pm0.003$ & $\mathbf{0.0226\pm0.002}$ & $\underline{0.0230\pm0.002}$ & $0.0270\pm0.002$ & $0.0269\pm0.003$ & $0.0273\pm0.003$ & $0.0226\pm0.002$ \\
    HR Analytic    & $0.0276\pm0.003$ & $\mathbf{0.0184\pm0.002}$ & $\underline{0.0187\pm0.002}$ & $0.0258\pm0.003$ & $0.0246\pm0.003$ & $0.0251\pm0.003$ & $0.0193\pm0.002$ \\
    German Credit  & $0.0640\pm0.007$ & $\mathbf{0.0549\pm0.006}$ & $\underline{0.0561\pm0.007}$ & $0.0615\pm0.006$ & $0.0601\pm0.007$ & $0.0631\pm0.010$ & $0.0565\pm0.006$ \\
    Breast Cancer  & $0.3949\pm0.017$ & $\mathbf{0.0429\pm0.009}$ & $\underline{0.1423\pm0.024}$ & $0.2805\pm0.032$ & $0.2701\pm0.043$ & $0.2256\pm0.073$ & $0.0474\pm0.006$ \\
    Loan           & $0.1413\pm0.015$ & $\mathbf{0.0803\pm0.014}$ & $\underline{0.1004\pm0.015}$ & $0.1275\pm0.016$ & $0.1122\pm0.028$ & $0.1105\pm0.020$ & $0.0763\pm0.014$ \\
            \midrule
            Rank & 6.0 & \textbf{1.0} & \underline{2.0} & 4.63 & 3.75 & 3.63 & $\emptyset$\\
        \bottomrule
    \end{tabular}
    \end{adjustbox}
    \label{tab:distribution_adaptation_metric}
\end{table}

\newpage
\section{AUROC per Iteration for GBS}
\begin{figure}[h]
    \centering
    \includegraphics[width=1.0\linewidth]{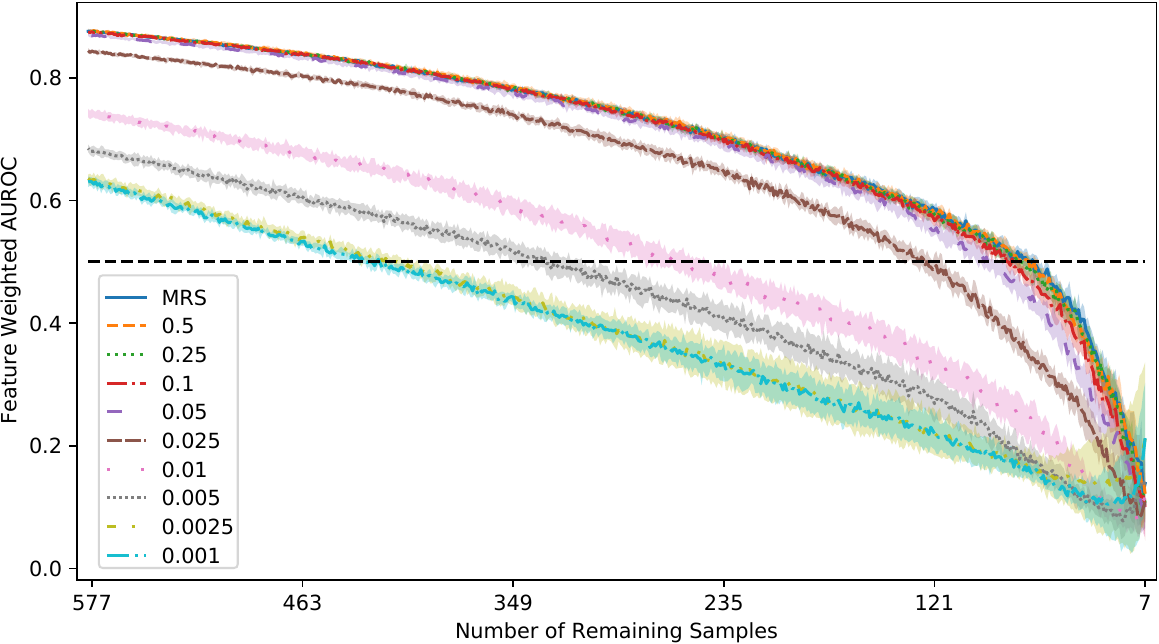}
    \caption{Comparison of mean AUROC of MRS vs. FW-MRS$_{RF}$ with different temperatures on 50 debiasing runs of GBS with auxiliary information of Allensbach.}
    \label{fig:gbs_allensbach_dropped_elements}
\end{figure}

This section examines the changes in AUROC over iterations and compares them across different temperatures. The mean and standard deviation for 50 repeated runs are shown. The diagram (\Cref{fig:gbs_allensbach_dropped_elements}) shows that fewer samples have to be dropped using FW-MRS$_{RF}$ compared to MRS. Using lower weights further reduces the number of dropped samples, starting slowly at the beginning and gradually increasing in strength later. The diagram illustrates the potential number of samples that can be retained by incorporating feature weights into the debiasing procedure.

\newpage
\section{Feature Importance GBS}
\begin{figure}[h!]
    \centering
    \includegraphics[width=1.0\linewidth]{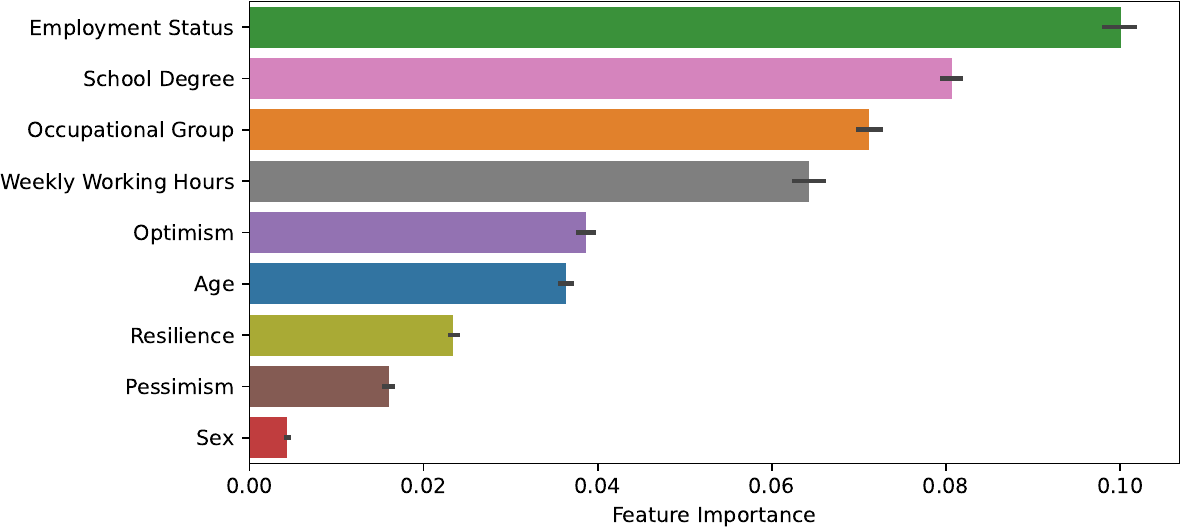}
    \caption{Feature importance used in FW-MRS$_{RF}$ debiasing of GBS with auxiliary information of Allensbach.}
    \label{fig:gbs_allensbach_important_features}
\end{figure}

\Cref{fig:gbs_allensbach_important_features} illustrates the feature importance of the unweighted domain classifier for GBS and Allensbach. It reveals that the most important and, hence, most biased features are associated with subjects' education and employment. This finding confirms the bias towards an overrepresentation of highly educated individuals.

\FloatBarrier

\end{document}